\documentclass[lettersize,journal]{IEEEtran}
\usepackage{amsmath,amsfonts}
\usepackage{algorithmic}
\usepackage{algorithm}
\usepackage{array}
\usepackage[caption=false,font=normalsize,labelfont=sf,textfont=sf]{subfig}
\usepackage{textcomp}
\usepackage{stfloats}
\usepackage{url}
\usepackage{verbatim}
\usepackage{graphicx}
\usepackage{cite}
\usepackage{ragged2e}
\usepackage{booktabs,makecell, multirow, tabularx}
\begin{document}

\title{Probabilistic Prediction of Longitudinal Trajectory\\Considering Driving Heterogeneity with \\ Interpretability}

\author{Shuli Wang, Kun Gao,~\IEEEmembership{Member,~IEEE}, Lanfang Zhang, Yang Liu, Lei Chen
\thanks{This work was supported by VINNOVA (ICV-safety, 2019-03418), the Area of Advance Transport and AI Center (CHAIR-CO-EAIVMS-2021-009) at Chalmers University of Technology, and the China Scholarship Council (CSC)
(202206260141). \textit{(Corresponding author: Kun Gao and Lanfang Zhang)}.}
\thanks{Shuli Wang is with the Key Laboratory of Road and Traffic Engineering, Ministry of Education, Tongji University, Shanghai 201804, China, with College of Transportation Engineering, Tongji University, 4800 Cao’an Highway, Shanghai, 201804, China, and also with Department of Architecture and Civil Engineering, Chalmers University of Technology,
Gothenburg, 41296, Sweden (e-mail: 1810984@tongji.edu.cn)}
\thanks{Kun Gao and Yang Liu are with the Department of Architecture and Civil Engineering, Chalmers University of Technology,
Gothenburg, 41296, Sweden (e-mail: gkun@chalmers.se)}
\thanks{Lanfang Zhang is with the Key Laboratory of Road and Traffic Engineering, Ministry of Education, Tongji University, Shanghai 201804, China, and also with the College of Transportation Engineering, Tongji University, 4800 Cao’an Highway, Shanghai, 201804, China (e-mail: zlf2276@tongji.edu.cn)}
\thanks{Lei Chen is with the RISE Research Institutes of Sweden, 417 56 Gothenburg,
Sweden (e-mail: lei.chen@ri.se)}
}



\maketitle
\begin{abstract}
Automated vehicles are envisioned to navigate safely in complex mixed-traffic scenarios alongside human-driven vehicles. To promise a high degree of safety, accurately predicting the maneuvers of surrounding vehicles and their future positions is a critical task and attracts much attention. However, most existing studies focused on reasoning about positional information based on objective historical trajectories without fully considering the heterogeneity of driving behaviors. Besides, previous works have focused more on improving models’ accuracy than investigating their interpretability to explore the extent to which a cause and effect can be observed within a system. Therefore, this study proposes a trajectory prediction framework that combines Mixture Density Networks (MDN) and considers the driving heterogeneity to provide probabilistic and personalized predictions. Specifically, based on a certain length of historical trajectory data, the situation-specific driving preferences of each driver are identified, where key driving behavior feature vectors are extracted to characterize heterogeneity in driving behavior among different drivers. With the inputs of the short-term historical trajectory data and key driving behavior feature vectors, a probabilistic LSTMMD-DBV model combined with LSTM-based encoder-decoder networks and MDN layers is utilized to carry out personalized predictions. Finally, the SHapley Additive exPlanations (SHAP) method is employed to interpret the trained model for predictions. The proposed framework is tested based on a wide-range vehicle trajectory dataset. The results indicate that the proposed model can generate probabilistic future trajectories with remarkably improved predictions compared to existing benchmark models. Moreover, the results confirm that the additional input of driving behavior feature vectors representing the heterogeneity of driving behavior could provide more information and thus contribute to improving the prediction accuracy.
\end{abstract}


\section{Introduction}
\IEEEPARstart{S}{afety} assurance has always been a top priority in the development of driving technologies such as advanced driver assistance systems (ADAS) and autonomous driving \cite{ref1}. To promise a high degree of safety, accurately predicting maneuvers of surrounding vehicles (SVs) and other traffic participants and their future positions is a critical task. Such a capability in various complex traffic scenarios is vital for autonomous vehicles (AVs) and assisted driving vehicles to plan safe and feasible maneuvers \cite{ref2}. In the future mixed traffic scenario with both autonomous vehicles and human-driving vehicles, it remains a huge challenge to accurately predict the trajectory of the surrounding human-driven vehicles because of the inherent heterogeneity in driving behavior and their vulnerability to the influence of surrounding traffic conditions. Due to the stochastic nature of human decision-making, probabilistic predictions are more reasonable that can produce possible future trajectories of the predicted vehicle \cite{ref3}, thus ensuring traffic safety more reliably with a higher probability. Therefore, probabilistic prediction is crucial for the safe driving of autonomous vehicles and receives much interest.

Probabilistic predictions of the trajectories of human-driven vehicles have a major difficulty in modeling the uncertainty of human driving behavior. Due to the intra-driver heterogeneity, the driver may make different driving decisions at different times in the same traffic scenario. For example, to avoid colliding with the preceding vehicle that suddenly decelerates, one can choose to decelerate at different rates, abruptly or slowly. To produce multiple possible future trajectories instead of one deterministic trajectory, existing studies have proposed some methods. Many scholars apply the Generative Adversarial Networks (GAN) \cite{ref4,ref5} and the Conditional Variational Auto-encoder (CVAE) \cite{ref6,ref7} to generate the distribution of the prediction. Additionally, some methods employ probability models to produce multiple potential trajectories along with their associated probabilities, such as Gaussian Mixture Model (GM) \cite{ref8} and Conditional Random Fields (CRF). However, most above methods utilize the prior Gaussian distribution to generate future trajectories, which is insufficient to cover the wide-spread space of the potential trajectories. Motivated by the work of Schwab, et al. \cite{ref9}, the Mixture Density Networks (MDN) that consist of multiple Gaussian distributions are used to combine with other deep learning algorithms to output distributions of arbitrary shapes, which could provide accurate probabilistic predictions.

Additionally, most current trajectory predictions reason about positional information based on short-term objective historical trajectories, without sufficiently considering the inherent stochasticity of driving behaviors caused by the inter-driver heterogeneity of human drivers. In fact, drivers are subjective and they always make different driving decisions even under the same traffic scenario. Early works tended to construct a trajectory prediction model considering different driving styles, which incorporates a module that first identifies driving styles \cite{ref10}. Also, some methods attempt to adopt the GAN model to generate different latent distributions for different drivers \cite{ref11,ref12}. However, this model didn’t always show good performance for the reason that the learned distribution is based on the short-term historical position information and sometimes can’t represent the personalized feature. Another approach is to construct a prediction model with the input of subjective personalized features, such as driving style labels and driving experience labels \cite{ref3}. Although the driving behavior characteristics are well considered in existing research, which and how driving behavior characteristic indicators can fully describe the heterogeneity of driving behavior and how to extract the driving behavior characteristics using realistic data in real application cases are still vague at present. Therefore, it is constructive to take driving behavior characteristics into consideration for establishing a personalized trajectory prediction model.

Moreover, the interpretability and traceability of the model is also a crucial aspect of trajectory predictions. For achieving accurate predictions, it is essential to be able to identify key influencing factors that affect the accuracy of predictions and infer what causes the future trajectories to vary among different samples. In addition, the model interpretability can help to prove whether driving behavior characteristics are effective inputs. However, a great majority of existing methods are unable to provide insight into the internal mechanism of models, especially deep learning-based methods.

This study aims to introduce a probabilistic trajectory prediction framework that combines MDN with the extensively used Long Short Term Memory (LSTM) encoder-decoder networks and takes into account the driver heterogeneity to provide personalized and explainable predictions. In the framework, the first module utilizes a certain length of historical trajectory to identify the situation-specific driving preferences of each driver and extract driving behavior feature vectors to characterize heterogeneity in driving behavior in the current driving scenario. These driving behavior feature vectors can help provide more information without greatly increasing the computational complexity of the prediction model. Subsequently, based on short-term historical trajectory data and the driving behavior feature vectors, the LSTMMD-DBV model is proposed for personalized trajectory prediction, where the encoder network is used to learn dynamic motion features of different samples and the MDN layer is used to provide probabilistic forecasting. Furthermore, the SHapley Additive exPlanations (SHAP) method is employed to provide explanations for predictions. The proposed framework is tested based on wide-range vehicle trajectory data containing trajectory data of all vehicles passing through a road section. The results demonstrate significant improvement in prediction accuracy and indicate that integrating additional driving behavior feature input contributes to improved trajectory prediction. 

The remainder of this paper is outlined as follows. Section \ref{Literature review} introduces the literature review mainly about trajectory prediction and interpretable algorithms. In Section \ref{Methodology}, the methodology for predicting the trajectory is proposed in detail, including representation of driving behavior characteristics, probabilistic trajectory prediction model, and interpretable method for predictions. A real-world trajectory dataset is collected, and the implementation details and comparison models of the proposed framework are described in Section \ref{Experiments}. Consequently, the results are presented and discussed in Sections \ref{Results} and \ref{Conclusions}.

\section{Literature review} \label{Literature review}
This section reviews existing studies on trajectory prediction and interpretable algorithms that are the most relevant to this study.

\subsection{Trajectory prediction} \label{Trajectory prediction}
Existing trajectory prediction methods can be generally divided into conventional mathematical methods \cite{ref13} and deep learning-based methods \cite{ref14}. Conventional mathematical methods build models based on expert-designed assumptions/rules that simplify the dynamic motion of vehicles and the interaction process between vehicles. Those models include physics-based models, maneuver-based models, and interaction-aware models. Despite the good interpretability of these models, their performances are usually compromised due to the inevitably introduced inappropriate artificial design.

Recently, a variety of works utilize deep learning techniques for trajectory prediction to achieve better predictive performance. The commonly-used deep neural network models address the prediction problem from two main aspects: most studies employed the Recurrent Neural Network (RNN) or its variants (e.g., Long Short-Term Memory (LSTM), Gated Recurrent Units (GRU), etc.) to capture temporal features from the historical trajectories \cite{ref15,ref16}; another branch of works apply Convolutional Neural Networks (CNN) \cite{ref17,ref18} and Graph Convolutional Networks (GCN) \cite{ref19,ref20} to encode spatial information from the surroundings. However, many existing models provide a deterministic output, which is not consistent with the indeterminacy of human behavior. 

To explore the uncertainty of future trajectories, many probabilistic prediction methods are proposed to address the multimodal output challenge. A portion of the method utilizes generative models, which incorporate a latent variable into predictors to produce different output samples corresponding to different output distributions, such as the GAN model \cite{ref15} and the CVAE model \cite{ref6}. However, there is a large performance gap in existing generative methods between the most-likely single output and the best posterior selection from distribution \cite{ref6,ref21}. The reason is that the pre-defined latent distribution does not learn individual driving behavior features well. Other approaches model multi-modal output distributions along with their associated probabilities by probability models, such as GM \cite{ref8}, MDN \cite{ref9}, and CRF. Among them, MDN is composed of multiple Gaussian distributions, which can fit any form of distribution and can model the uncertainty of human behavior well.

Due to the limitations in data availability and computational power, most existing trajectory prediction models consider short-term historical trajectory information (a few seconds to a few minutes of trajectory data) as input, without sufficiently taking the subjective heterogeneity of drivers into account. Some studies forecast the future states of vehicles based on a few seconds of vehicle trajectory data \cite{ref22}. Additionally, some researches also include other surrounding traffic environments (e.g., traffic conditions, road geometry, traffic signals, etc.) to improve the prediction accuracy \cite{ref23,ref24}. However, the abovementioned inputs of short-term trajectory and surrounding traffic environment provide little information about the driving behavior characteristics. To address this problem, some researchers simply construct a trajectory prediction model considering different driving styles, which incorporates a module that first identifies driving styles \cite{ref10}. Some methods attempt to adopt the GAN model to generate different latent distributions for different drivers \cite{ref12}. Another approach is to incorporate subjective personalized features into the prediction model, such as driving style labels and driving experience labels \cite{ref3}. Although the driving behavior characteristics are considered in existing studies, the precise characterization of driving behavior characteristics based on realistic data in real application scenarios is not clear enough and needs further research. For instance, some studies used driving styles as one of the features in trajectory prediction and simply assumed that the driver’s driving style is prior information obtained from other places, which was not realistically feasible in real applications.

\subsection{Interpretable algorithms} \label{Interpretable algorithms}
One major limitation of deep learning models is that their prediction process is often a “black box” that cannot be demonstrated and visualized for human understanding. Consequently, the study of interpretable algorithms has become increasingly soaring in recent years \cite{ref25}.

Currently, interpretable algorithms adopted to explain trained deep learning models can be mainly categorized into global interpretability and local interpretability \cite{ref26}. Global interpretability refers to explaining the internal working mechanism behind models, providing an overview of the relationships between the input features and the output predictions. Local interpretability aims to help understand the decision process of models for a specific instance, providing insight into how the model makes its decision on a case-by-case basis. For prediction problems, this paper mainly focuses on local interpretability approaches.

A few mainstream interpretable algorithms have been applied to explain deep learning-based models for applications of trajectory prediction \cite{ref27}, including Local Interpretable Model-agnostic Explanations (LIME), Partial Dependence Plots (PDP), and SHAP. The LIME method is a local interpretable algorithm, which has the advantage of not requiring knowledge of the specific model type or any modifications of the model itself \cite{ref28}. However, it has the disadvantage of high computational cost for high-dimensional data. Besides, PDP can visualize the relationship between a target variable and one or more predictor variables while holding other predictors constant \cite{ref29}. However, PDP is not well-suited for interpreting LSTM-based models since PDPs cannot capture the complex temporal dynamics modeled by LSTMs. Compared to LIME, the SHAP method is capable of handling high-dimensional data and takes into account the interactions between features \cite{ref30}. What’s more, the SHAP method has merits that it can provide both local and global interpretation. Therefore, this study investigates the explainability of the trajectory prediction model based on the SHAP method.

\section{Methodology} \label{Methodology}
\begin{figure}[!t]
\centering
\includegraphics[width=3.45in]{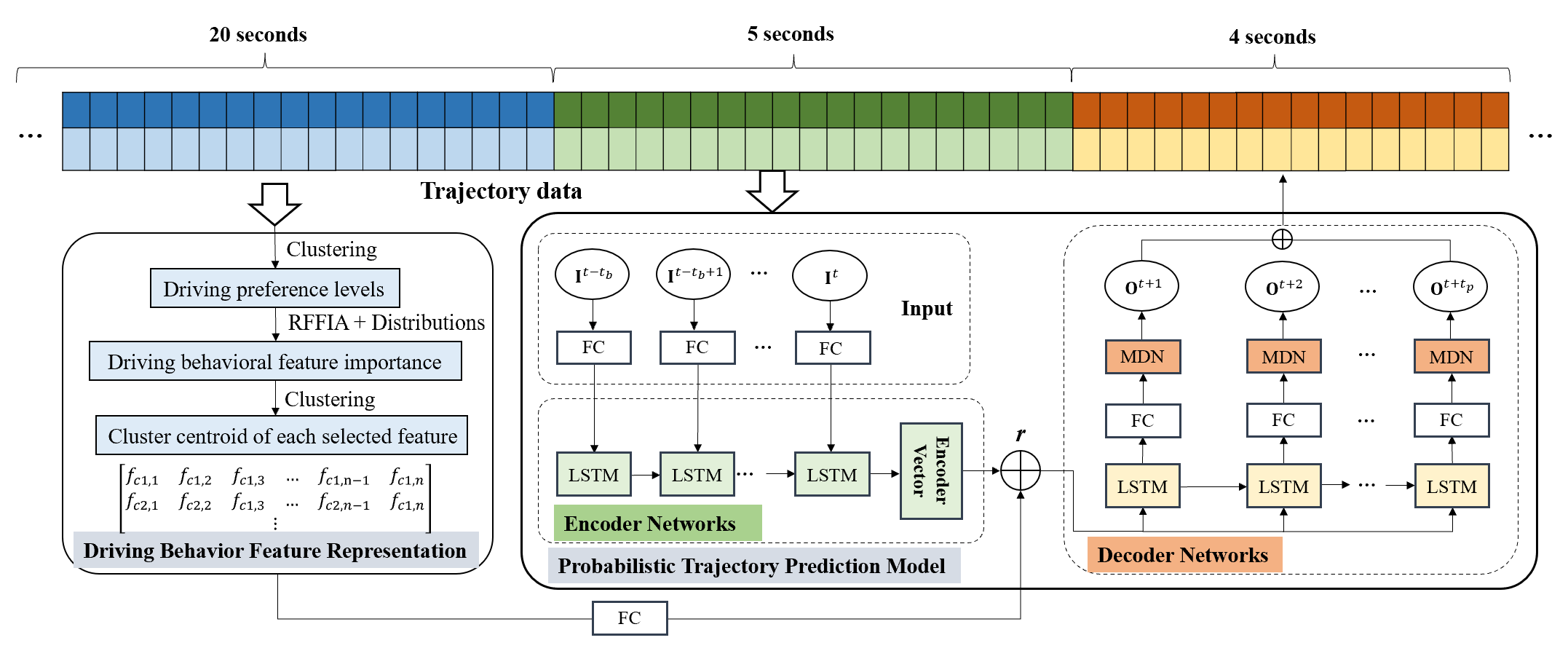}
\caption{The overall architecture of trajectory prediction considering driving heterogeneity.}
\label{fig:1}
\end{figure}

This section describes the detailed methodologies of the proposed model. Fig. \ref{fig:1} illustrates the overall architecture of the proposed longitudinal trajectory prediction model considering driving heterogeneity. The overall architecture can be divided into two parts. First, 20-second historical trajectory information before the starting point of prediction is utilized to quantify the driving preferences of each driver, where driving behavior feature vectors are extracted to characterize heterogeneity in driving behavior among different drivers. We select 20-second historical trajectory information as such inputs are feasible to be obtained in real applications through intelligent transport systems such as video monitors and Vehicle-to-everything (V2X), which ensures the applicability of the proposed model in reality. Second, the LSTM-based encoder-decoder networks combining MDNs are developed and trained to predict the longitudinal trajectory using historical dynamic motion information and the driving behavior feature vectors. The input to the LSTM-based encoder networks is the five-second trajectory of subjective and surrounding vehicles. Detailly, an encoder network containing LSTM cells takes in the historical trajectory data of five seconds to extract the time-sequence patterns of trajectories to generate vehicle motion feature vectors. Then the driving behavior feature vectors and motion feature vectors are concatenated and inputted into the decoder network. Lastly, the LSTM-based decoder networks output the distribution of future longitudinal trajectories for the next four seconds. By considering the driving behavior characteristics of each vehicle, the connected LSTM module is trained to achieve personalized trajectory prediction. More importantly, using the MDN layer instead of a typical linear output layer could contribute to predicting the probability estimates of a vehicle’s possible future trajectories. In addition, a post-hoc method of interpretation SHAP is employed to provide explanations for predictions of the proposed model. The following section explains each of the modules in detail.

First, the method of driving behavior feature representation is introduced in section \ref{3.1} to accurately explore the heterogeneity of driver behavior. Section \ref{3.2} elaborates on the specific details of the probabilistic trajectory prediction model based on the LSTMMD-DBV network, including the encoder module, decoder module, and mixture density network module. Finally, section \ref{3.3} introduces the interpretable algorithm SHAP method to interpret the predictions of the probabilistic trajectory prediction model.

\subsection{Representation of Driving Behavior Characteristics} \label{3.1}
As one major contribution of this study, situation-specific driving preferences are identified to accurately characterize the heterogeneity in driving behavior among different drivers. Most of the previous works focus on investigating driving behavior characteristics by identifying drivers' long-term driving styles using questionnaire surveys or vehicle operation data for one trip, several days, or even several months. However, these approaches face challenges related to driver information privacy, data acquisition difficulties, and substantial computational requirements in practical applications. In addition, a driver’s driving style is unstable and may change his/her previous driving style when the driving scenarios change. Therefore, this study quantifies the situation-specific driving preference of each driver based on 20-second historical trajectory information before the starting point of prediction, to more accurately extract and characterize heterogeneity in driving behavior in the current driving scenarios. The historical trajectory data with short observation windows, which is applied to identify the driving preference and served as one of the inputs for the corresponding trajectory prediction, can be feasibly obtained in real applications through intelligent transportation systems such as video monitors and V2X. This ensures the applicability of the proposed model for in-vehicle control and real-time predictions.

This section introduces how to identify the situation-specific driving preferences of drivers and extract the driving behavior characteristic vectors to characterize heterogeneity in driving behavior among different drivers in the current driving scenarios. Feature indicators are firstly extracted based on a certain length of historical trajectory data. Then driving preference levels are determined by the dimensionality reduction method and clustering method. The Feature Importance Analysis method is applied to screening out key feature indicators. Finally, the clustering method is adopted to determine the number of categories for each selected feature indicator, where cluster centroid values instead of continuous values and clustering categories are used to describe the differences in the feature indicator among different drivers.

\subsubsection*{\bf Driving Behavior Feature Indicators}
To better characterize the heterogeneity of driving behavior, this study applies three types of feature extraction methods, i.e., time-domain features, frequency-domain features, and sequence-domain features, as explained as follows.

(1) Time-domain features can be extracted by commonly used statistical methods (SM), e.g., the maximum, minimum, mean, and variance of variables, which can describe most of the distribution information of the data. What’s more, the volatility in driving is utilized to measure the variations of instantaneous driving decisions in recent studies and is proven to be effective in characterizing driving behavior \cite{ref31}. Two metrics of volatility feature, i.e., Mean Absolute Deviation and Time-Varying Stochastic Volatility, are calculated as Eqs. \eqref{eq:1} - \eqref{eq:2}.

Mean Absolute Deviation (MAD) quantifies the average distance between the observation of a variable and its central tendency:

\begin{equation}
\label{eq:1}
\text{MAD} = \frac{\sum_{i=1}^{n}\mid x_{i}-\overline{x} \mid}{n}.
\end{equation}
where $x_i$ denotes the $i_{th}$ value of the feature $x$, $\overline{x}$ denotes the mean of the observed feature $x$, and $n$ is the sample size. 

Time-varying Stochastic Volatility (TSV) measures the fluctuation of data by computing the changes in the proportion of observations:

\begin{subequations}\label{eq:2}
\begin{align}
\text{TSV} &= \sqrt{\frac{\sum_{i=1}^{n} (r_{i}-\overline{r})^2}{n-1}} \label{eq:2A}\\
r_{i} &= ln(\frac{x_i}{x_{i-1}}) \times 100 \label{eq:2B}
\end{align}
\end{subequations}
where $x_i$ and $x_{i-1}$ are the $i_{th}$ and $(i-1)_{th}$ value of the feature, respectively, and $\overline{r}$ is the mean of $r_i$.

(2) Frequency-domain features are captured by discrete wavelet transform (DWT) and discrete fourier transform (DFT). These two methods convert the time series variables into signal amplitudes in the frequency domain. In this study, the indicators gravity center of frequency (GCF), root mean square of frequency (RMSF), mean square frequency (MSF), and standard deviation of frequency (STDF) is applied in the DFT method, while indicators wavelet energy entropy (WEE) and wavelet singular entropy (WSE) are applied in the DWT method. The details of these methods can be found in \cite{ref32,ref33}.

(3) Sequence-domain features are utilized to capture the diversity and heterogeneity of different driving trajectory sequences. Weighted dynamic time warping (WDTW) is a powerful tool for evaluating the similarity between given sequences by matching the corresponding points in the two sequences and calculating the distance between them \cite{ref34}. This study uses WDTW to measure the similarity between trajectories (velocity and acceleration). 

In this study, vehicle velocity, acceleration, and deceleration data are calculated by the above methods, i.e. SM , DWT, DFT , and WDTW to select the most suitable method to facilitate the driving preference classification model, as shown in Table \ref{tab:table1}. Besides, time headway is mainly employed to represent the traffic flow while time to collision (TTC)  is used to indicate the occurrences of dangerous events in car-following situations.

\begin{table*}[!t]
\caption{The driving behavior feature indicators \label{tab:table1}}
\centering
\begin{tabular}{c c c l c}
\hline
\rule{0pt}{8pt}
\textbf{Features} & \textbf{Units} & \textbf{Method} & \textbf{Indicators} & \textbf{Symbol} \\
\hline
\rule{0pt}{8pt}
\multirow{13}{*}{Velocity} & \multirow{13}{*}{m/s} & \multirow{6}{*}{SM} & Maximum value of the velocity sequence & $MAX_v$ \\
 & & & Minimum value of the velocity sequence & $MIN_v$ \\
 & & & Mean value of the velocity sequence & $MEAN_v$ \\
 & & & Variance of the velocity sequence & $VAR_v$ \\
 & & & Mean Absolute Deviation (MAD) of the velocity sequence & $MAD_v$ \\
 & & & Time-varying Stochastic Volatility (TSV) of the velocity sequence & $TSV_v$ \\
 \cline{3-5}
 \rule{0pt}{8pt}
 & & \multirow{4}{*}{DWT} & Gravity center of frequency (GCF) of the velocity sequence & $GCF_v$ \\
 & & & Root mean square of frequency (RMSF) of the velocity sequence & $RMSF_v$ \\
 & & & Mean square frequency (MSF) of the velocity sequence & $MSF_v$ \\
 & & & Standard deviation of frequency (STDF) of the velocity sequence & $STDF_v$ \\
 \cline{3-5}
 \rule{0pt}{8pt}
 & & \multirow{3}{*}{DFT} & Gravity center of frequency (GCF) of the velocity sequence & $GCF_v$ \\
 & & & Wavelet energy entropy (WEE) of the velocity sequence & $WEE_v$ \\  
 & & & Wavelet singular entropy (WSE) of the velocity sequence & $WSE_v$ \\  
 \cline{3-5}
 \rule{0pt}{8pt}
 & & WDTW & Similarity between two velocity sequences & $WDTW_v$ \\
\hline
\rule{0pt}{8pt}
\multirow{15}{*}{Acceleration} & \multirow{15}{*}{m/s$^2$} & \multirow{8}{*}{SM} & Maximum value of the acceleration sequence & $MAX_a$ \\
 & & & Minimum value of the acceleration sequence & $MIN_a$ \\
 & & & Maximum value of the deceleration sequence & $MAX_d$ \\
 & & & Minimum value of the deceleration sequence & $MIN_d$ \\
 & & & Mean value of the acceleration sequence & $MEAN_a$ \\
 & & & Mean value of the deceleration sequence & $MEAN_d$ \\
 & & & Variance of the acceleration sequence & $VAR_a$ \\
 & & & MAD of the acceleration sequence & $MAD_a$ \\
 \cline{3-5}
 \rule{0pt}{8pt}
 & & \multirow{4}{*}{DWT} & GCF of the acceleration sequence & $GCF_a$ \\
 & & & RMSF of the acceleration sequence & $RMSF_a$ \\
 & & & MSF of the acceleration sequence & $MSF_a$ \\
 & & & STDF of the acceleration sequence & $STDF_a$ \\
 \cline{3-5}
 \rule{0pt}{8pt}
 & & \multirow{3}{*}{DFT} & GCF of the acceleration sequence & $GCF_a$ \\
 & & & WEE of the acceleration sequence & $WEE_a$ \\ 
 & & & WSE of the acceleration sequence & $WSE_a$ \\  
\hline
\rule{0pt}{8pt}
THW & s & SM & Minimum value of time headway & $MIN_{thw}$ \\
\hline
\rule{0pt}{8pt}
TTC & s & SM & Minimum value of time to collision & $MIN_{ttc}$ \\
\hline
\end{tabular}
\end{table*}

\subsubsection*{\bf Feature Indicators Dimensionality Reduction and Visualization}
It is technically challenging to cluster multivariate data with specific driving preferences based on the above-mentioned large number of feature indicators. Driving preferences refer to a driver’s behavioral characteristics in a specific driving scenario, reflecting the driver’s different preferences to make behavioral choices in different driving scenarios. Meanwhile, features depicting driving behaviors that do not vary across different drivers make few contributions to differentiating driving preferences, so such features should be omitted. To reduce the dimensionality of the dataset before clustering, the t-SNE algorithm is adopted to project high-dimensional features onto two-dimensional space, which can easily distinguish the differences between clusters \cite{ref35}. Besides, the t-SNE algorithm can visualize the features, which could show the extracted features’ capacity for distinguishing driving preferences.

\subsubsection*{\bf Driving Preferences Identification by Clustering Algorithm}
To cluster driving preferences into several types, unsupervised learning clustering method \textit{k}-medoids is utilized to extract driving preferences from large amounts of features. The \textit{k}-medoids method is a clustering algorithm similar to the well-known \textit{k}-means method \cite{ref36}. Different from the \textit{k}-means uses the mean of the cluster, the \textit{k}-medoids method applies the medoids of the cluster that are the actual member points to update the cluster centroid, which makes it more robust to outliers without being affected by outliers. Moreover, as the cluster centers in \textit{k}-medoids are actual data points, each cluster's center represents the most representative point within that cluster, allowing for a direct understanding of the characteristics of each cluster. The criterion function for selecting a medoid of the cluster is that the sum of the distances from the remaining member points in the current cluster to that medoid is minimized. Due to this difference, the \textit{k}-medoids method is more robust to noise and outliers in the dataset. To determine the number of clusters \textit{K}, a popular internal evaluation metric i.e., the Silhouette Coefficient (SC) is employed. The SC values range from -1 to +1, with high values indicating a great cluster configuration. Finally, the number of driving preference categories and the labels (driving preference categories) for each trajectory are obtained.

\subsubsection*{\bf Driving Behavior Feature Indicators}
Based on the extracted feature indicators and driving preference categories, the Random Forest Feature Importance Analysis (RFFIA) method is adopted to rank the feature indicators that influence driving preference classification \cite{ref37}. The RFFIA method obtains the feature importance based on the Random Forrest classification model, which regards the average contribution of each feature to each decision tree in the random forest as the importance of each feature. In this study, the “ \textit{feature importances} ” function of the random forest model in the Scikit-learn Python library is called for obtaining feature importance scores of features. The contribution of each feature is computed by the “mean decrease in impurity” mechanism called Gini importance. Then, each feature is ranked according to its importance score to get a list of the most important features. Besides, the distribution of each feature indicator is checked. Feature indicators that fall within a small range should not be considered because they do not show significant differences between samples and therefore cannot express sample diversity. Finally, the driving behavior feature vector for the trajectory of a vehicle can be represented as
\begin{equation}
\label{eq:3}
\mathbf{D}_i = [{f_{i,1},f_{i,2},...,f_{i,k}}]
\end{equation}
where $f_{i,k}$ denotes the $k_{th}$ feature of the $i_{th}$ trajectory.

For each selected feature indicator, the cluster centroid values are used to describe the differences in the feature indicator among different drivers. The original feature indicator values are not adopted, as continuous variables cannot express the differences between samples, such as the maximum acceleration value, where there is no significant difference between 1.0 m/s$^\text{2}$ and 1.1 m/s$^\text{2}$. Meanwhile, we don’t directly apply the categorical labels of each feature indicator as inputs to the model. The reason is that, compared to categorical labels, cluster centroid values may have more physical significance and can more accurately express the differences between samples. Taking the minimum speed indicator as an example, compared to inputting label one and label two, cluster centroid values 16.5 m/s$^\text{2}$ and 21.6 m/s$^\text{2}$ may better express significant differences in driving preferences for speed among drivers. To determine the cluster centroid values, the clustering method is employed and the number of clustering categories is determined when the SC value reaches its maximum.

\subsection{Probabilistic Trajectory Prediction Model by LSTMMD-DBV Network} \label{3.2}
The trajectory prediction problem can be regarded as a time series modeling problem. The main task is to predict the future trajectory of the target vehicle (TV) based on historical vehicle motion information and driving behavior characteristics of the TV and surrounding vehicles (SVs). The majority of studies confirmed that LSTMs are particularly efficient for time series prediction because they can remember previous inputs, and have been widely applied for predicting vehicle trajectory. However, LSTMs are deterministic in nature and output deterministic values of future acceleration or position. Therefore, to achieve probabilistic prediction of a vehicle’s possible future trajectories, this study proposed an LSTMMD-DBV network that combines LSTM-based encoder-decoder networks and MDN dense layers.

The prediction model proposed in this section is displayed in  Fig. 1. First, vehicle motion characteristic variables are given as input to the encoder module of the prediction model. They are represented as
\begin{equation}
\label{eq:4}
\mathbf{I}_{m,i}^t = [\mathbf{m}_{i}^{t-t_b},\mathbf{m}_{i}^{t-t_b+1},\ldots,\mathbf{m}_{i}^{t}]
\end{equation}
where
\begin{equation}
\label{eq:5}
\mathbf{m}^t = [{\Delta v_{1}}^{t},{\Delta v_{2}}^{t},\ldots,{\Delta v_{n}}^{t},{\Delta x_{1}}^{t},{\Delta x_{2}}^{t},\ldots,{\Delta x_{n}}^{t},{v}^{t},{a}^{t}]
\end{equation}
where $\Delta v_n^t$ and $\Delta x_n^t$ represent the relative velocity and the relative distance between the TV and all the SVs in the longitudinal direction at time $t$, respectively. The subscript $n$ denotes the ID of the SVs. We consider six vehicles within 150m in the longitudinal direction and within two adjacent lanes. $v^t$ and $a^t$ represent the instantaneous velocity and acceleration of the TV in the longitudinal direction at time $t$, respectively. $t_b$ denotes the observation horizon.

Since driving behavior feature vectors aren’t time-series data, key driving behavior feature vectors would be embedded into the same dimension as the vehicle motion characteristic vectors. Then, the embedded driving behavior feature vector is merged with the encoder vector of the vehicle motion feature for each trajectory, resulting in the trajectory encoding vector. The final trajectory encoding vector inputted into the decoder module of the prediction model is displayed as
\begin{equation}
\label{eq:6}
\mathbf{R}_{i}^t = Concat(\mathbf{EV}_{i}^t, \mathbf{ED}_i)
\end{equation}
where $\mathbf{EV}_{i}^t$ is the encoder vector of the vehicle motion feature of the $i_{th}$ trajectory at time $t$. $\mathbf{ED}_i$ is the embedded driving behavior feature vector of the $i_{th}$ trajectory, i.e., $\mathbf{ED}_i=\text{Emb}(\mathbf{D}_i)$. $Concat()$ is a function to merge vectors.

The output of the trajectory prediction model is velocity distribution, which is denoted as
\begin{equation}
\label{eq:7}
\mathbf{O}_i^t = [\mathbf{p}_i^{t+1}, \mathbf{p}_i^{t+2}, \ldots, \mathbf{p}_i^{t+t_p}]
\end{equation}
where $\mathbf{p}_i^{t}$ is the predicted velocity distribution of the $i_{th}$ trajectory at each time step $t$ and $t_p$ is the prediction horizon.

\subsubsection*{\bf LSTM Encoder}
The encoder network is designed as four stacked LSTM layers to learn and capture the dynamic patterns of historical information for each trajectory. The historical trajectory information of the TV and its surrounding vehicles is encoded into a fixed-length motion characteristic vector. This vector contains the understanding and memory regarding the features of historical trajectories.

In our LSTM encoder implementation, the inputted feature vector is first embedded by a fully connected (FC) layer and fed into the LSTM encoder. The initial hidden state of the encoder network is set to a null vector. The LSTM encoder updates the current hidden state based on the current embedded input vector and the hidden state from the previous time step. The equation of the LSTM encoder can be expressed as
\begin{equation}
\label{eq:8}
\mathbf{h}_e^t = LSTM_{ec}(FC(\mathbf{I}_m^t),\mathbf{h}_e^{t-1})
\end{equation}
where $FC()$ is a fully connected layer embedding the input features into a higher space. $LSTM_{ec}()$ means the shared LSTM encoder used in this model. $\mathbf{h}_{e}^t$ is the hidden state of the encoder at time $t$.

Subsequently, the hidden state of the LSTM network is processed by an FC layer to obtain the encoder vector as the final representation of the dynamic features of the vehicle motion. Usually, the final hidden state of the encoder that summarizes all information in the input historical trajectory is used as the initial state for the decoder. To output the hidden state as the encoder vector that can be merged with the driving behavior feature vector, the FC layer is employed as follows
\begin{equation}
\label{eq:9}
\mathbf{EV}^t = FC(\mathbf{h}_e^t) = f(\mathbf{w^T} \cdot \mathbf{h}_e^t + \mathbf{b})
\end{equation}
where $\mathbf{w}$ denotes the weight matrix and $\mathbf{b}$ means the bias vector.

\subsubsection*{\bf LSTM Decoder}
In the prediction model framework, an LSTM decoder is applied to generate the probabilistic future trajectory of the TV. Specifically, to incorporate the heterogeneity of driving behavior into the output of the decoder for trajectory prediction, the encoder vector of the vehicle motion feature and the embedded driving behavior feature vector are concatenated and inputted into the LSTM decoder. In the decoding stage, the LSTM decoder updates the current hidden state based on the output driving behavior vector, the hidden state from the previous time step as well as the trajectory encoding vector. The current output vector is updated based on the current hidden state, the output vector of the previous moment, and the trajectory encoding vector. The equation of the LSTM decoder is
\begin{align}
\mathbf{h}_d^t &= LSTM_{dc}(\mathbf{h}_d^{t-1},\mathbf{q}^{t-1},\mathbf{R}) \label{eq:10} \\
\mathbf{q}^t &= g(\mathbf{h}_d^t,\mathbf{q}^{t-1},\mathbf{R}) = sigmoid(\mathbf{h}_d^t,\mathbf{q}^{t-1},\mathbf{R}) \label{eq:11}
\end{align}
where $LSTM_{dc}()$ is the LSTM decoder used in this model. $\mathbf{h}_{d}^t$ is the hidden state of the decoder at time $t$. $\mathbf{q}^t$ denotes the output vector at time $t$. $sigmoid()$is an S-shaped function commonly used as an activation function in neural networks to map variables to a range between 0 and 1.

Next, the output vector of the LSTM decoder is passing through an FC layer to adjust its dimensionality to obtain $z^t$ that aligns with MDN. Finally, to capture the inherent uncertainty of drivers’ decision-making, the MDN is employed to outcome the distribution of the future velocity of the TV. The MDN structure will be introduced in detail in the next section. The following equation shows the LSTM decoder:
\begin{align}
\mathbf{z}^t &= FC(\mathbf{q}^t) \label{eq:12} \\
\mathbf{O}^t &= MDN(\mathbf{z}^t) \label{eq:13}
\end{align}
where $\mathbf{z}^t$ denotes an intermediate process vector and $MDN()$ is the MDN layer.

\subsubsection*{\bf Mixture Density Networks}
The MDNs are neural networks that can be connected to the last layer of the neural network to output possible distributions of predicted values. Generally, the MDN models the distribution as a mixture of multiple Gaussian distributions, and outputs the parameters of each Gaussian (mean, standard deviation, and mixture weight coefficient), as shown in Eqs. \eqref{eq:14} - \eqref{eq:15}. Theoretically, the MDN can approximate any conditional probability distribution.

\begin{align}
P(\mathbf{O}|\mathbf{z}) &= \sum_{c=1}^C \pi_c(\mathbf{z}) D(\mathbf{y}|\mu_c(\mathbf{z}), \sigma_c(\mathbf{z}))
 \label{eq:14} \\
D(\mathbf{O}|\mu_c(\mathbf{z}), \sigma_c(\mathbf{z})) &= \frac{1}{\sqrt{2\pi\sigma_c(\mathbf{z})}}\exp\left(-\frac{[\mathbf{z}-\mu_c(\mathbf{z})]^2}{2\sigma_c^2(\mathbf{z})}\right)
 \label{eq:15}
\end{align}
where $\mathbf{O}$ is the output and $\mathbf{z}$ is the input to the MDN. $C$ is the number of mixtures. $\pi_c(\mathbf{z})$ denotes the mixture weight and $D(\mathbf{O}|\mu_c(\mathbf{z}),\sigma_c(\mathbf{z}))$ represents a multidimensional Gaussian distribution function consisting of two parameters, i.e., the mean values $\mu_c(\mathbf{z})$ and the standard deviation $\sigma_c(\mathbf{z})$.

The MDN outputs a different probability distribution at each prediction time step. Given input vectors to the MDN to generate probability distributions for $t_f$ prediction horizon, the constraints and generation of parameters are as follows:
\begin{subequations}\label{eq:16}
\begin{align}
\pi_c(\mathbf{z}) &= Softmax(\mathbf{z}) \label{eq:16A}\\
\mu_c(\mathbf{z}) &= Linear(\mathbf{z})  \label{eq:16B} \\
\sum_c(\mathbf{z}) &= exp(\mathbf{z})  \label{eq:16C}
\end{align}
\end{subequations}
where $\pi_c(\mathbf{z})$ is a set of $C$ values obtained through SoftMax activation to ensure their sum to one, i.e., $\sum_{c=1}^C \pi_c (\mathbf{z}) = 1$; $\mu_c (\mathbf{z})$ is a $C \times t_f$ estimated by a simple linear function; $\sum_c(\mathbf{z})$ is a $C \times t_f$ obtained through exponential function.

The probabilistic prediction model combining the LSTM-based networks and MDNs considers the uncertainty of prediction results. Therefore, the negative log-likelihood is applied as the objective function train the proposed model:
\begin{equation}
\label{eq:17}
L_{NLL} = -\frac{1}{N} \sum_{i\in N} \log(P(\mathbf{O}_i|\mathbf{I}_i))
\end{equation}
where $N$ is the number of samples. $\mathbf{I}_i$ and $\mathbf{O}_i$ are the input and output of the network, respectively.

\subsection{Interpreting Predictions with SHAP Method} \label{3.3}
This section intends to explore the influence of the input feature variables on the prediction results, especially the driving behavior features. SHAP is a post-hoc method of interpretation, which explains an opaque model by measuring the importance of each feature in the results predicted by the model, making it applicable to any model \cite{ref30}. In this study, the SHAP method is applied when the proposed LSTMMD-DBV model makes a prediction, to estimate the influence level of each feature on the predicted outcome, and identify them as positive or negative. In addition, this study can reveal whether and to what extent the driving behavior feature indicators contribute to accurate trajectory prediction. 

The SHAP method is an additive feature attribution method that aims to approximate the full and complex model by using simpler and interpretable models. The core idea of the SHAP method is to treat all features as “contributors” and calculate the marginal contributions of each feature to the model output, which are named Shapley values. Given a specific input x, the simplified explanation model h is shown as:
\begin{equation}
\label{eq:18}
f(x) = h(x') = \Phi_0 + \sum_{i=0}^M \Phi_i x_i'
\end{equation}
where $f()$ is the original model, $h()$ is the simplified model, and $x_i'$ is one of $M$ simplified features. $\Phi_0$ is a constant value with all simplified inputs toggled off. $\Phi_i$ represents the contribution of feature $i$, which is measured by a Shapley value and can be calculated as follows:
\begin{equation}
\label{eq:19}
\Phi_i = \sum_{S\subseteq F-\{i\}} \frac{|S|!(F-|S|-1)!}{F!} [f_x(S\cup\{i\}) - f_x(S)]
\end{equation}
where $F$ is the set of all input features. $S\subseteq F-\{i\}$ means the subset of all features excluding feature $i$. $|S|$ denotes the number of non-zero features in $S$. $f_x(S\cup\{i\})$ represents the overall contribution of all features including the $i_{th}$ feature, so the difference $[f_x(S\cup\{i\}) - f_x(S)]$ represents the contribution of the $i_{th}$ feature on the prediction result.

\section{Experiments} \label{Experiments}
\subsection{Data Preparation and Processing} \label{4.1}
To validate the effectiveness of the proposed vehicle trajectory prediction model, real-world vehicle trajectory data collected from the TJRD TS platform (available online at https://tjrdts.com) is used \cite{ref38}. The dataset contains 7 days of trajectory data of all vehicles passing through a highway section (1.5 km in length) located in Hangzhou, China. The road segment has three lanes and an emergency lane on each side, of which the width of the lane is 3.75 m and the width of the emergency lane is 3.5 m. The data is collected by millimeter wave radars installed on both sides of the road section with a collection frequency of 10 HZ per second. The original trajectory database contains all frame-series-based data for each track, such as the initial time and the final time of the tracks, the timestamp, the longitudinal and lateral position (Lane ID) information, the longitudinal and lateral velocity, the actual lateral offset, and physical features such as the type of the vehicle.

\begin{figure}[!t]
\centering
\includegraphics[width=3.0in]{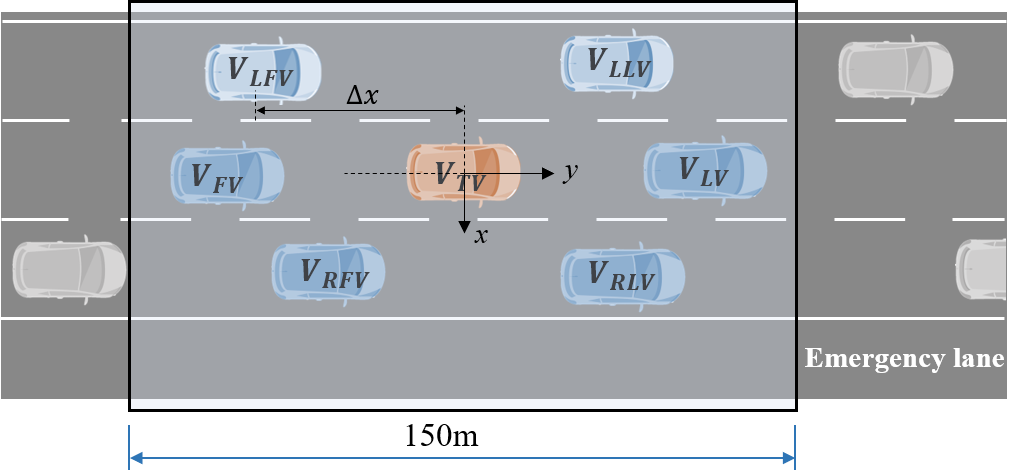}
\caption{Schematic of the TV and its surroundings.}
\label{fig_2}
\end{figure}

Due to the obstruction of other objects and the measurement error of the equipment, the original data is preprocessed. Firstly, abnormal track records with a total number of tracks below 100 are deleted, missing timestamp data are interpolated, and duplicate timestamps are deleted. Secondly, actual lateral offset data and longitudinal velocity are repaired and smoothed. The longitudinal acceleration of the vehicle is obtained by differential calculation of the speed profile. Then, SVs are matched according to the Lane ID and their longitudinal position. Taking the perceptive range of on-board sensors into account, the most influential six SVs within 150 m in the longitudinal direction and within two adjacent lanes are selected. The position relationships of the TV and its surroundings are shown in Fig. \ref{fig_2}. In the schematic diagram, LV, RLV, and LLV represent the leading vehicle in the current lane, the right adjacent lane, and the left adjacent lane, and FV, RFV, and LFV represent the following vehicle in the current lane, the right adjacent lane, and the left adjacent lane, respectively. The position and velocity of the surrounding vehicles are converted into the form of the difference from the TV. If there is no vehicle at the location within the specified range, the position and velocity information of the vehicle at this location is set to \(\Delta x_s^t = +\infty\) and \(v_s^t = v_{\text{TV}}^t\). Finally, more than 2,697 trajectories of more than 50 seconds are extracted from the original dataset and applied to train and validate the proposed trajectory prediction model.

\subsection{Implementation Details} \label{4.2}
The hyperparameters of the proposed model and all the comparison models are optimized by using a grid search technique on train and validation sets. The train and validation data are created by cross-validation with a split ratio of 4:1. The trajectory set is large enough to have a considerable number of samples (more than 240,000) for the train and validation sets. There are more than 193,000 samples for training and 48,000 samples for validation at each training iteration. The optimized hyper-parameters for the proposed trajectory prediction model are displayed in \ref{tab:table2} and the hyper-parameters of the comparison models are explained in Section 4.3. Observation length and prediction horizons are 5s and 4s, respectively. The LSTM cell consists of three layers of stacked hyper LSTM recurrent units with a dropout rate of 0.4. The optimizers adopt the ADAM optimizer.

\begin{table}[!t]
\caption{The optimized hyper-parameters for the proposed trajectory prediction model \label{tab:table2}}
\centering
\begin{tabular}{c c}
\hline
\rule{0pt}{8pt}
\textbf{Hyperparameters} & \textbf{Settings} \\
\hline
\rule{0pt}{8pt}
Learning rate & 0.001 \\
Encoder length & 50 \\
Decoder length & 40 \\
Number of neurons & 192 \\
Dropout rate & 0.2 \\
Batch size & 100 \\
Epochs & 50 \\
Number of Mixtures & 5 \\
\hline
\end{tabular}
\end{table}

\subsection{Models Comparison} \label{4.3}
We compare the prediction results of our proposed model with several benchmark models that have been employed in relevant studies. The following is the list of comparison models along with a brief description of each one.
\begin{itemize}
\item{Gaussian process (GP)}: a Gaussian Process Regression model is trained to predict the trajectory. The implementation of GP is carried out using the scikit-learn library in Python, with the historical trajectories of the TV and SVs.
\item{LSTM}: the model has the same LSTM cell as the proposed model, which consists of two layers of stacked hyper LSTM recurrent units. 
\item{LSTMGM}: the model has the same LSTM cell as the proposed model, with the combination of the Gaussian Mixture (GM) model as the output layer. 
\item{LSTMMD}: the encoder and decoder networks are similar to the proposed model, with the input feature of vehicle motion characteristics. The tuned hyper-parameters for this model consist of: (1) optimizer: Adam optimizer, (2) batch size: 100, (3) learning rate: 0.001, and (4) the number of mixtures: 4, which are all tuned by a grid search.
\item{LSTMMD-DP}: the model has the same model framework as the proposed model, with the input feature of driving preference labels besides vehicle motion characteristics. The tuned hyper-parameters for this model consist of: (1) optimizer: Adam optimizer, (2) batch size: 100, (3) learning rate: 0.001, and (4) the number of mixtures: 4, which are all tuned by a grid search. 
\end{itemize}

To accurately assess the performance of the proposed model and its benchmarks, two evaluation metrics are employed: Root Mean Squared Error (RMSE) and Root-Weighted Square Error (RWSE).

1) RMSE: a commonly used evaluation metric for assessing prediction performance. The equation is shown as
\begin{equation}
\label{eq:20}
\text{RMSE} = \sqrt{\frac{1}{t_p} \sum_{i=1}^{t_p} (v_H^{(i)} - v)^2}
\end{equation}
where $v_H^{(i)}$ denotes the ground-true value in the $i_th$ trajectory at time horizon $H$. $t_p$ represents the prediction horizon of the output samples. $a$ is the $i_th$ prediction result out of the output samples. In fact, RMSE calculates the error between predicted values and true values to quantify the accuracy of the model. For probabilistic prediction results, the RMSE is calculated by taking the mean of the probability distribution at each time step.

2) RWSE: a reasonable evaluation metric for probability distribution outputs, which captures the deviation of a model’s probability mass from real-world trajectories, and is calculated as follows
\begin{equation}
\label{eq:21}
\text{RWSE} = \sqrt{\frac{1}{t_p} \sum_{i=1}^{t_p} \int_{-\infty}^{\infty} p(v) (v_H^{(i)} - v)^2 dv}
\end{equation}
where $p(v)$ is the modeled density of velocity. $m$ denotes the sample size of the test set used for model evaluation. To reduce the complexity of the calculation, we estimated it with $n$=500 simulated traces per recorded trajectory by the Monte Carlo method based on the following equation
\begin{equation}
\label{eq:22}
\text{RWSE} = \sqrt{\frac{1}{t_p \times n} \sum_{i=1}^{t_p} \sum_{j=1}^{n} (v_H^{(i)} - \hat{v}_H^{(i,j)})^2}
\end{equation}
where $\hat{v}_H^{(i,j)}$ represents the simulated velocity result under sample $j$ for the $i_th$ trajectory at time horizon $H$.

\section{Results} \label{Results}
\subsection{Key Driving Behavior Feature Vectors} \label{5.1}
\begin{figure}[!t]
\centering
\subfloat[]{\includegraphics[width=2.5in]{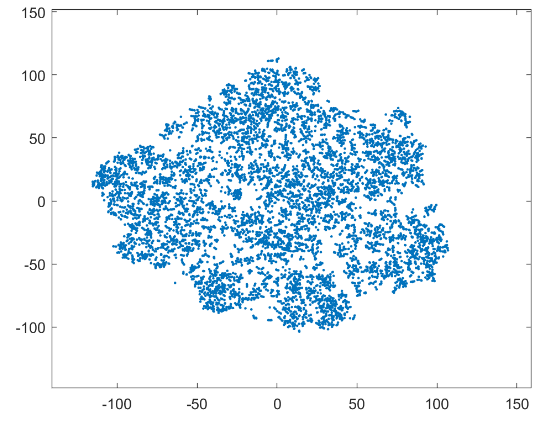}%
\label{fig3_first_case}} \\
\subfloat[]{\includegraphics[width=2.45in]{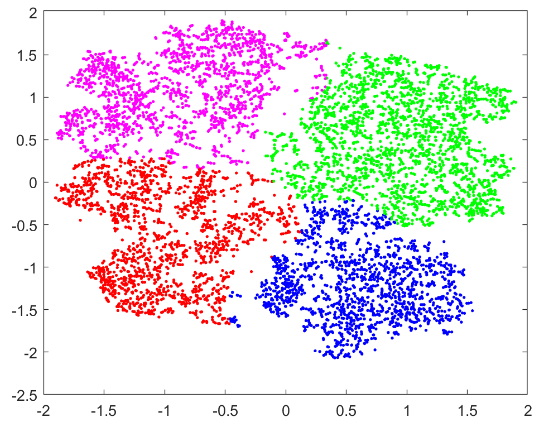}%
\label{fig3_second_case}}
\caption{Features distribution of PCA and t-SNE methods in the driving preference classification model. (a) PCA. (b) t-SNE.}
\label{fig_3}
\end{figure}

Based on the low-dimensional features after t-SNE dimensionality reduction, the driving preferences are clustered into four categories where the SC value reaches its maximum. The dimension reduction result with driving behavior features of the driving preferences classification model is shown in Fig. \ref{fig_3}. As we can see, the combination of the t-SNE and k-medoids method can separate driving preferences well. Compared with PCA, the feature distribution with the t-SNE method is significantly more separable.

\begin{figure}[!t]
\centering
\includegraphics[width=3.2in]{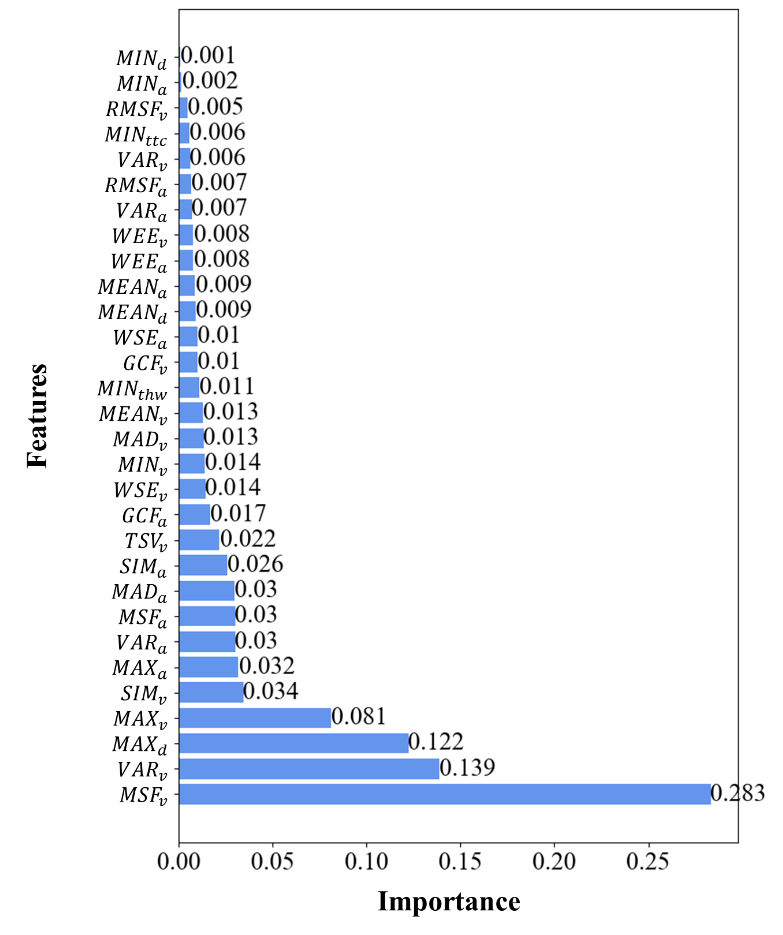}
\caption{The feature importance of driving behavior feature indicators.}
\label{fig_4}
\end{figure}

Subsequently, the feature importance of driving behavior feature indicators is obtained by the RFFIA method and displayed in Fig. \ref{fig_4}. As can be seen, the indicator $MSF_v$ has the largest feature importance with a value of 0.2835, implying the frequency of the velocity sequence has a significant effect on the type of driving preferences, i.e., rapid changes in velocity play a key role. In addition, indicators including $VAR_v$, $MAX_d$, and $MAX_v$ also have significant impacts on determining driving preferences. This study selects a cumulative feature importance threshold of 90\% to filter out key driving behavior feature indicators, to reach a balance between reducing the number of features and retaining sufficient information.

\begin{figure*}[!t]
\centering
\includegraphics[width=6in]{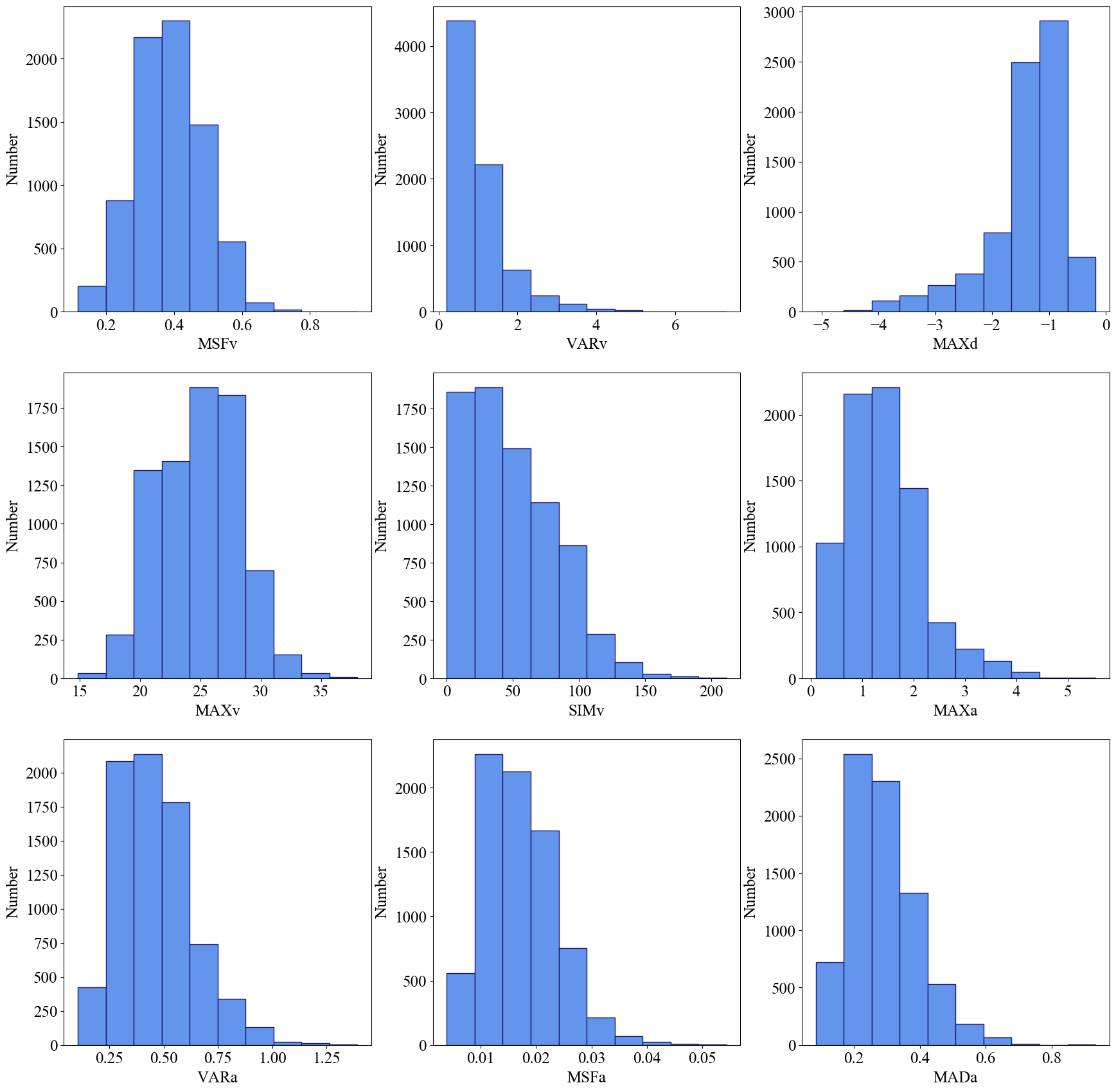}
\caption{The distributions of driving behavior feature indicators.}
\label{fig_5}
\end{figure*}

The distribution of each driving behavior feature indicator is checked. For example, Fig. \ref{fig_5} shows the distribution of the top nine feature indicators of the feature importance list. As we can see, distributions of eight feature indicators are within a relatively large range, except for the feature indicator $MSF_a$ which is distributed in a smaller range of [0, 0.06],  meaning that this indicator does not differ noticeably among different samples. Therefore, the indicator $MSF_a$ is not considered in the key driving behavior feature list. The remaining feature indicators are also checked according to the above process. For each selected feature indicator, the cluster centroid values are used to describe the differences in the feature indicator among different drivers.

To determine the optimal dimension of key driving behavior feature vectors, the performance of the proposed model with the input of different dimensional feature vectors is tested. Based on the feature importance result, the input feature dimensions range from 4 to 16. The result illustrates that the performance of the proposed model initially improves with an increase in the input feature dimension, and then reaches a stable state. The optimal dimension of key feature vectors is 7 where the performance of the model reaches its best.

\subsection{Performance Comparison of Different Prediction Models} \label{5.2}
With a prediction horizon of 4s, Table \ref{tab:table3} reports the velocity prediction results in terms of RMSE and RWSE of the LSTMMD-DBV model and other benchmark models. It can be observed that the proposed model achieves the best performance in terms of RWSE when compared with the other models. Specifically, the proposed model achieves an average improvement of 3.8\% in the prediction error of velocity compared with the LSTMMD model. The GP model performs the worst since the temporal dynamics of the vehicle motion are barely taken into account. Meanwhile, the LSTM model performs better than the GP model, because the LSTM model is capable of learning the temporal variability of vehicle motions.

Moreover, compared with the LSTM model, the LSTMGM model and the LSTMMD model achieve pretty large improvements in velocity predictions by outputting probabilistic future velocities. That is because, in dynamic traffic environments, the future trajectory of the vehicle may not be fully deterministic due to complex road environments and users. Notably, the LSTMMD model performs much better than the LSTMGM model, implying that the MDN is a better model for future velocity predictions than the GM. In addition, compared with the LSTMMD model, the LSTMMD-DP model achieves a more superior performance in velocity predictions due to taking into account the driving preferences of drivers. Furthermore, the LSTMMD-DBV model performs better than the LSTMMD-DP model due to considering driving behavior characteristics vectors rather than category labels of driving preferences. It can be inferred that taking driving behavior feature vectors into account could provide more information and has a positive effect on improving the accuracy of the model prediction, which corroborates the meaningful contribution of the proposed model.

\begin{table}[!t]
\caption{Comparison of velocity prediction results \label{tab:table3}}
\centering
\begin{tabular}{l c c}
\hline
\rule{0pt}{8pt}
\textbf{Models} & \textbf{RMSE (m/s)} & \textbf{RWSE (m/s)} \\
\hline
\rule{0pt}{8pt}
GP & 1.339 & 1.340 \\
LSTM & 1.283 & 1.282 \\
LSTMGM & 1.258 & 1.255 \\
LSTMMD & 1.242 & 1.238 \\
LSTMMD-DP & 1.231 & 1.226 \\
LSTMMD-DBV (proposed model) & \textbf{1.195} & \textbf{1.189} \\
\hline
\end{tabular}
\end{table}

\subsection{Evaluation of Trajectory Prediction Model} \label{5.3}
\begin{figure*}[!t]
\centering
\includegraphics[width=6.6in]{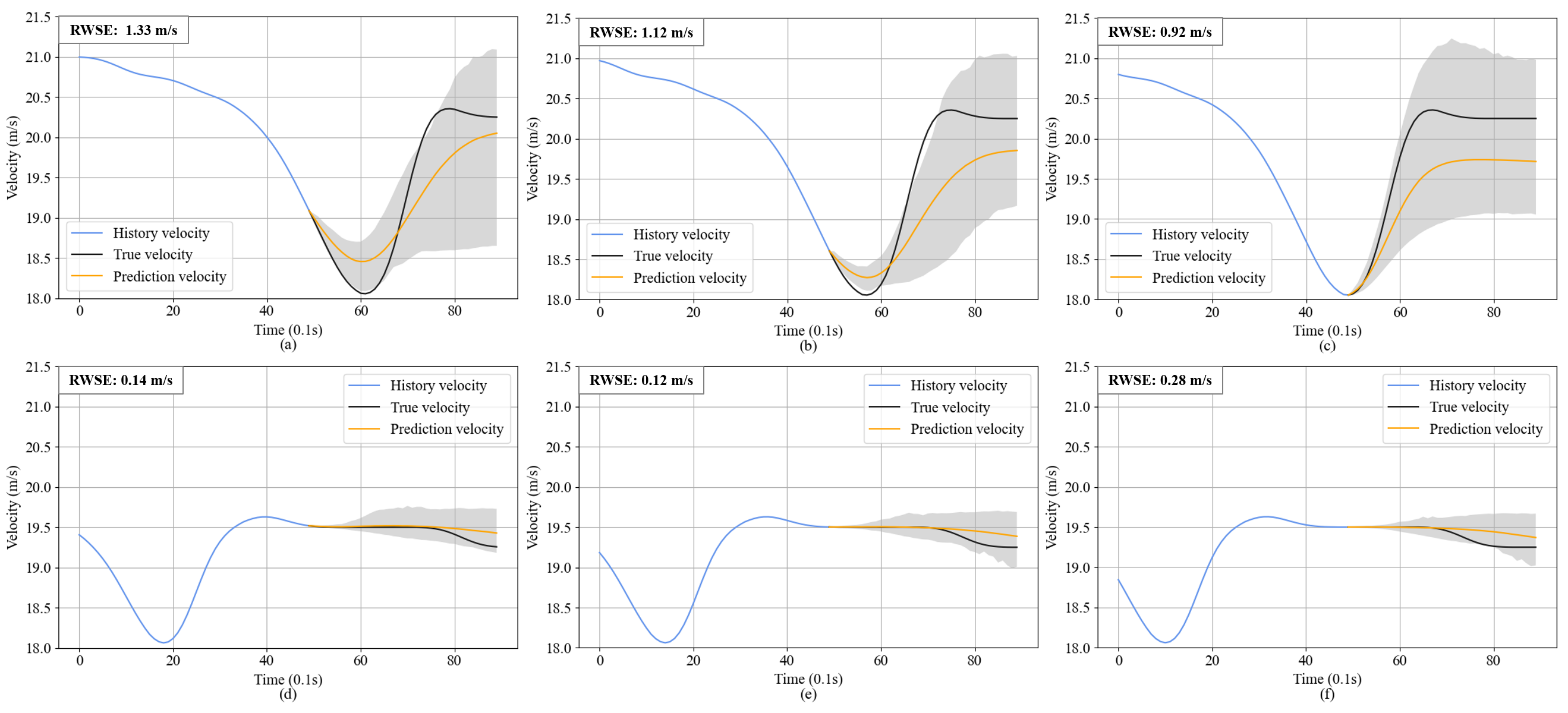}
\caption{The velocity prediction results under different traffic scenarios.}
\label{fig_6}
\end{figure*}

\begin{figure*}[!t]
\centering
\includegraphics[width=6.6in]{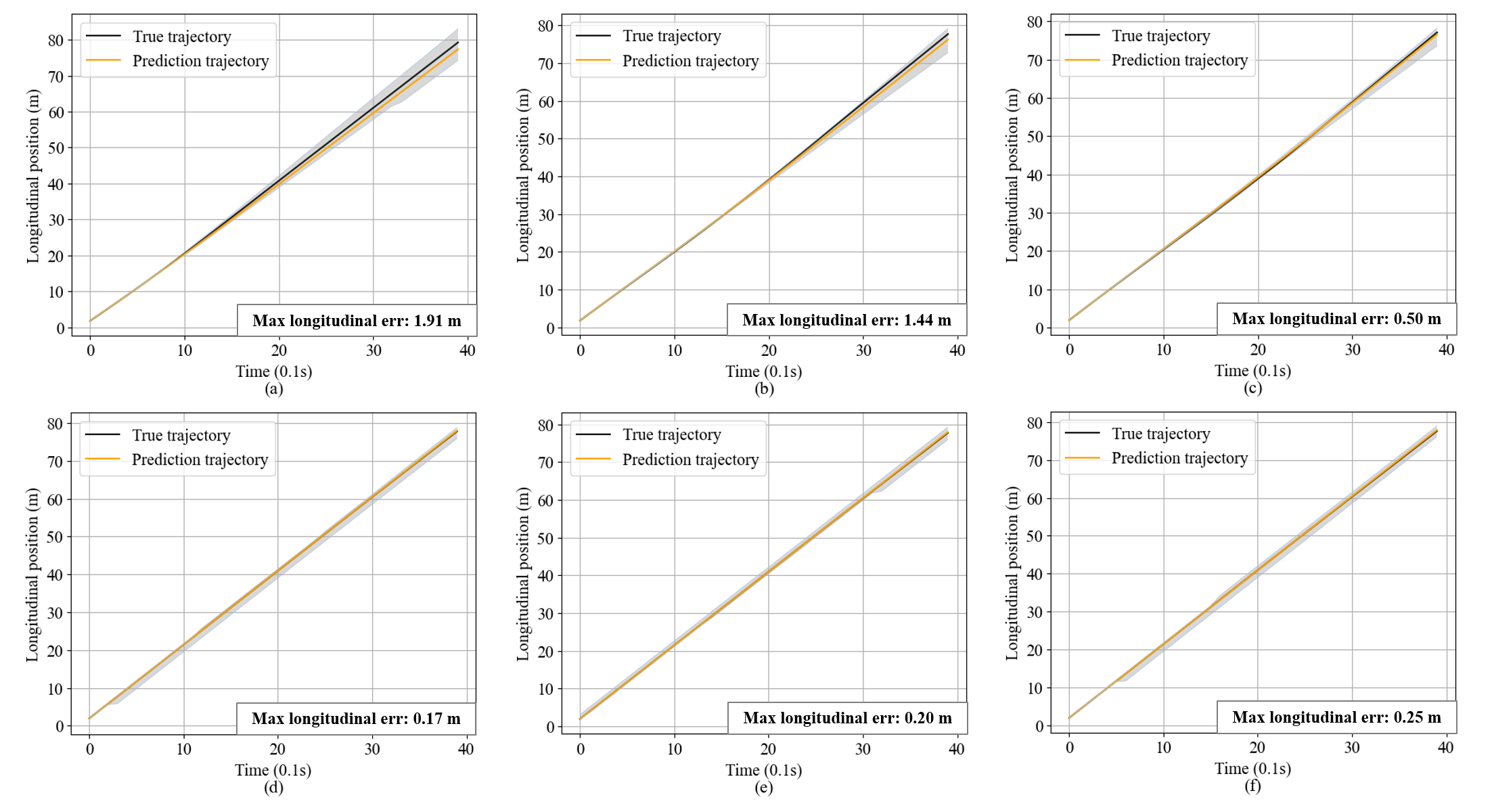}
\caption{The longitudinal trajectory prediction results under different traffic scenarios.}
\label{fig_7}
\end{figure*}

To intuitively understand trajectory prediction results, Fig. \ref{fig_6} and Fig. \ref{fig_7} illustrate the velocity prediction and trajectory prediction results under different traffic scenarios. The subplots labeled as (a) - (c) in the Fig. \ref{fig_6} and Fig. \ref{fig_7} demonstrate a scenario with significant speed fluctuations while subplots labeled as (d) - (f) demonstrate a scenario with stable speed. In Fig. \ref{fig_6}, the blue line denotes the history velocity sequence containing the 5s data in total, which is used as input to the proposed trajectory prediction model. The black line denotes the ground-truth trajectory. The proposed trajectory prediction model generates prediction results for the next 4s at each time step, which is shown by the orange line in Fig. \ref{fig_6} and Fig. \ref{fig_7}. Meanwhile, the gray broadband represents the 95\% confidence interval of the predicted probability distribution, which displays the possible trajectories. Besides, only the position error in the longitudinal direction at the last timestamp is displayed to investigate the prediction accuracy. Clearly, it can be observed in Fig. \ref{fig_6} and Fig. \ref{fig_7} that the ground-truth values of velocity and position mostly fall within the 95\% confidence intervals of the velocity and position prediction results, which indicates that the proposed prediction model can generally cover the potential future trajectories.

In the scenario where the TV accelerates and decelerates, the velocity prediction results in Fig. \ref{fig_6} (a) - (c) indicate that the prediction model doesn’t perform well. When the predicted starting time is far from the moment of speed change, as shown in Fig. 6 (a), the RWSE of the predicted velocity is 1.33 m/s, which demonstrates bad predictive performance. As the predicted starting time approaches the moment of speed change, the RWSE decreases. Fig. \ref{fig_6} (c) shows that when the predicted starting time is close to the moment of speed change, the RWSE of the predicted velocity decreases to 0.92 m/s, and the prediction model can distinguish whether the vehicle will continue to decelerate or accelerate. Similar trends can be found for the corresponding prediction results of longitudinal position. As shown in Fig. \ref{fig_7} (a) - (c), the longitudinal position error decreases as the predicted starting time approaches the moment of speed change.

In the case where the TV remains at a steady velocity, the velocity and position prediction results in Fig. \ref{fig_6} (a) - (c)  and Fig. \ref{fig_7} (a) - (c) denote that the prediction model exhibits good predictive performance. It can be observed that the RWSE of the predicted velocity is 0.14 m/s, 0.12 m/s, and 0.28 m/s, respectively, which illustrates that the prediction model performs well for stable velocity sequences. Fig. \ref{fig_7} (c) shows that the longitudinal position error increases compared with Fig. \ref{fig_7} (a), where the longitudinal position error is 0.17 m. This can be explained by the fact that the inputs sent more information about driving with a stable speed to the prediction model, making it difficult for the model to perceive upcoming speed reductions.

\subsection{Analysis of the Influencing Factors in Predictions} \label{5.4}
\begin{figure}[!t]
\centering
\includegraphics[width=3.3in]{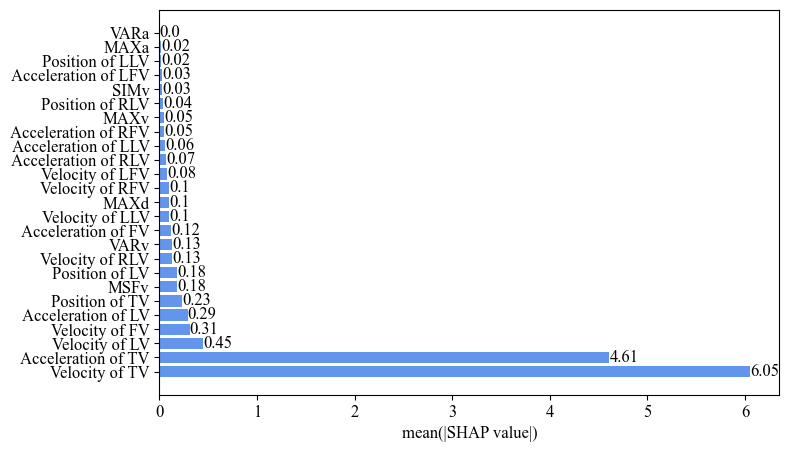}
\caption{The SHAP results of the proposed model.}
\label{fig_8}
\end{figure}

As mentioned, the SHAP method is used to learn the feature importance and how each features contributes to the generation of predicted values by the model, and the results are shown in Fig. \ref{fig_8}. As can be seen, a clear indication is that the historical velocity and acceleration of the target vehicle play a dominant role in the velocity predictions, which is consistent with the fact that the temporal features of the predicted vehicle's historical velocity are the basis for predicting its future trajectories. The velocity and acceleration of the leading vehicle in the same lane are also important factors, which is consistent with the classic car-following theory that these factors serve as stimuli for the response of the following vehicle.

As also observed, some driving behavior feature indicators contribute to the proposed model prediction. The $MSF_v$ and $VAR_v$ that represent the volatility of the velocity are important for the prediction process, potentially due to their ability to provide information on the driver's preference for driving stability. If the driver performs low velocity fluctuations and maintains a relatively stable velocity during the driving process, it may indicate a preference for stable driving. The $MAX_d$ that describes the deceleration operational characteristics of the driver could assist in providing driver's deceleration preference for the prediction process. Besides, the $MAX_v$, $SIM_v$, and $MAX_a$ are also beneficial for improving prediction performance.

\section{Conclusions} \label{Conclusions}
This study proposes an LSTMMD-DBV model that incorporates the heterogeneity of driving behavior to generate probabilistic and personalized trajectory predictions. The driving behavior feature vector representing the heterogeneity of driving behavior is taken into account to provide subjective information about driving behaviors and preferences of different drivers for trajectory prediction. The outputs of the trajectory prediction model are generated by MDNs to achieve accurate probabilistic predictions. Moreover, the SHAP method is employed to explain which are the important features and what role these features play in determining future driving behavior, to provide interpretability for the trained model. The proposed models are tested and validated on samples extracted from a wide-range vehicle trajectory dataset in a highway section. 

In the proposed framework, driving behavior feature indicators are extracted by three types of feature extraction methods, i.e., time-domain features, frequency-domain features, and sequence-domain features. For each selected feature indicator, instead of continuous values and clustering labels, the cluster centroid values are used to describe the differences in the feature indicator among different drivers. 

The performance comparison results show that the LSTMMD-DBV model outperforms the LSTMMD and the LSTMMD-DP model by a desirable improvement of the trajectory prediction accuracy. The results demonstrate that, by taking into account driving behavior feature vectors consisting of the cluster centroids of the selected feature indicators, more information is provided for model predictions, and a positive effect on improving model prediction is achieved. Besides, compared with the conventional LSTM model with deterministic predictions, the LSTMMD model achieves a better performance due to generating probabilistic predictions.

From the model interpretability perspective, the SHAP results illustrate that the historical velocity and acceleration of the target vehicle play a dominant role in the velocity predictions. As for the driving behavior feature indicators, the $MSF_v$ and $VAR_v$ that represent the volatility of the velocity and provide information on the driver's preference for the driving stability for model prediction are also shown to be important for the prediction process.

This study focuses on the inputs and outputs of the prediction framework to construct a probabilistic trajectory prediction model that considers the heterogeneity of driving behavior while little improvement has been done on the algorithms of the prediction model itself. Meanwhile, only short-term longitudinal trajectories are considered. As the next steps, the work can be extended to long-term two-dimensional trajectory prediction using more advanced and remarkable prediction algorithms, which could provide longer reaction time for trajectory planning and decision-making of autonomous vehicles to ensure safe driving. Besides, the proposed model should be tested on more driving scenarios to improve the generalization performance of the model.

\section*{Acknowledgments}
The authors are grateful to VINNOVA (ICV-safety, 2019-03418), the Area of Advance Transport and AI Center (CHAIR-CO-EAIVMS-2021-009) at Chalmers University of Technology, the China Scholarship Council (CSC) (202206260141) for funding this research.

\bibliographystyle{IEEEtran}
\bibliography{Trajectory_prediction.bib}

\end{document}